\documentclass[letterpaper]{article} 
\usepackage{aaai24}  
\usepackage{times}  
\usepackage{helvet}  
\usepackage{courier}  
\usepackage[hyphens]{url}  
\usepackage{graphicx} 
\usepackage{natbib}  
\usepackage{caption} 
\usepackage{algorithm}
\usepackage{amsmath,amsfonts}
\usepackage{algorithm}
\usepackage{array}
\usepackage[caption=false,font=normalsize,labelfont=sf,textfont=sf]{subfig}
\usepackage{textcomp}
\usepackage{url}
\usepackage{verbatim}
\usepackage{graphicx}
\usepackage{cite}
\usepackage{times}
\usepackage{epsfig}
\usepackage{graphicx}
\usepackage{amsmath}
\usepackage{amssymb}
\usepackage{url}
\usepackage{array}
\usepackage{multirow}
\usepackage{adjustbox}
\usepackage{threeparttable}
\usepackage{algorithm}
\usepackage[table,xcdraw]{xcolor}
\usepackage{amsmath}
\usepackage{amssymb}
\usepackage{multirow}
\usepackage{adjustbox}
\usepackage{threeparttable}
\usepackage{soul}
\usepackage{natbib} 
\usepackage{graphicx}
\usepackage{amsmath}
\usepackage{amssymb}
\usepackage{booktabs}
\usepackage{breqn}
\usepackage{subcaption}
\usepackage{tikz}
\usepackage{pgfplots}
\usepgfplotslibrary[groupplots]

\usepackage{makecell}

\usepackage{caption}
\usepackage{multirow}
\usepackage{subcaption}
\usepackage{booktabs}
\usepackage{bm}
\usepackage{algorithm, algpseudocode}
\usepackage{color}
\usepackage{colortbl}
\usepackage{textcomp}
\usepackage[outercaption]{sidecap}
\usepackage{makecell}
\usepackage{stmaryrd}
\usepackage{caption}
\usepackage{enumitem}
\usepackage{array}
\usepackage{enumitem}
\usepackage{pifont}
\usepackage{sidecap}
\usepackage{xspace}

\newcommand{\cmark}{\ding{51}}%
\newcommand{\xmark}{\ding{55}}%

\definecolor{Gray}{gray}{0.90}
\definecolor{LightCyan}{rgb}{0.82,0.82,1}

\newcommand{\name}{M$^2$-CLIP~}
\newcommand{\nameo}{M$^2$-CLIP}
\usepackage{newfloat}
\usepackage{listings}
\DeclareCaptionStyle{ruled}{labelfont=normalfont,labelsep=colon,strut=off} 
\floatstyle{ruled}
\newfloat{listing}{tb}{lst}{}
\floatname{listing}{Listing}
\title{M$^2$-CLIP: A Multimodal, Multi-task Adapting Framework \\  for Video Action Recognition}
\author{
   Mengmeng Wang\textsuperscript{\rm 1},  Jiazheng Xing\textsuperscript{\rm 1}, Boyuan Jiang\textsuperscript{\rm 2},\; Jun Chen\textsuperscript{\rm 1}, \\Jianbiao Mei\textsuperscript{\rm 1}, Xingxing Zuo\textsuperscript{\rm 3}$^*$, Guang Dai\textsuperscript{\rm 4}, Jingdong Wang\textsuperscript{\rm 5}, Yong Liu\textsuperscript{\rm 1}\thanks{Corresponding authors: Yong Liu and Xingxing Zuo.}
}
\affiliations{
    \textsuperscript{\rm 1}Zhejiang University \\ \textsuperscript{\rm 2}Youtu Lab,Tencent \\ \textsuperscript{\rm 3}Technical University of Munich \\ \textsuperscript{\rm 4}SGIT AI Lab, State Grid Corporation of China \\  \textsuperscript{\rm 5}Baidu Inc\\
 \{mengmengwang, jiazhengxing, junc, jianbiaomei\}@zju.edu.cn, byronjiang@tencent.com \\
xingxing.zuo@tum.de, guang.gdai@gmail.com, wangjingdong@outlook.com, yongliu@iipc.zju.edu.cn
}
\begin{document}

\maketitle

\begin{abstract}
Recently, the rise of large-scale vision-language pretrained models like CLIP, coupled with the technology of Parameter-Efficient FineTuning (PEFT), has captured substantial attraction in video action recognition. Nevertheless, prevailing approaches tend to prioritize strong supervised performance at the expense of compromising the models' generalization capabilities during transfer. In this paper, we introduce a novel Multimodal, Multi-task CLIP adapting framework named \name to address these challenges, preserving both high supervised performance and robust transferability.
Firstly, to enhance the individual modality architectures, we introduce multimodal adapters to both the visual and text branches. Specifically, we design a novel visual TED-Adapter, that performs global Temporal Enhancement and local temporal Difference modeling to improve the temporal representation capabilities of the visual encoder. Moreover, we adopt text encoder adapters to strengthen the learning of semantic label information.
Secondly, we design a multi-task decoder with a rich set of supervisory signals to adeptly satisfy the need for strong supervised performance and generalization within a multimodal framework.
Experimental results validate the efficacy of our approach, demonstrating exceptional performance in supervised learning while maintaining strong generalization in zero-shot scenarios.
\end{abstract}

\section{Introduction}
   \begin{figure}[t]
    \centering
    \includegraphics[width=.45\textwidth]{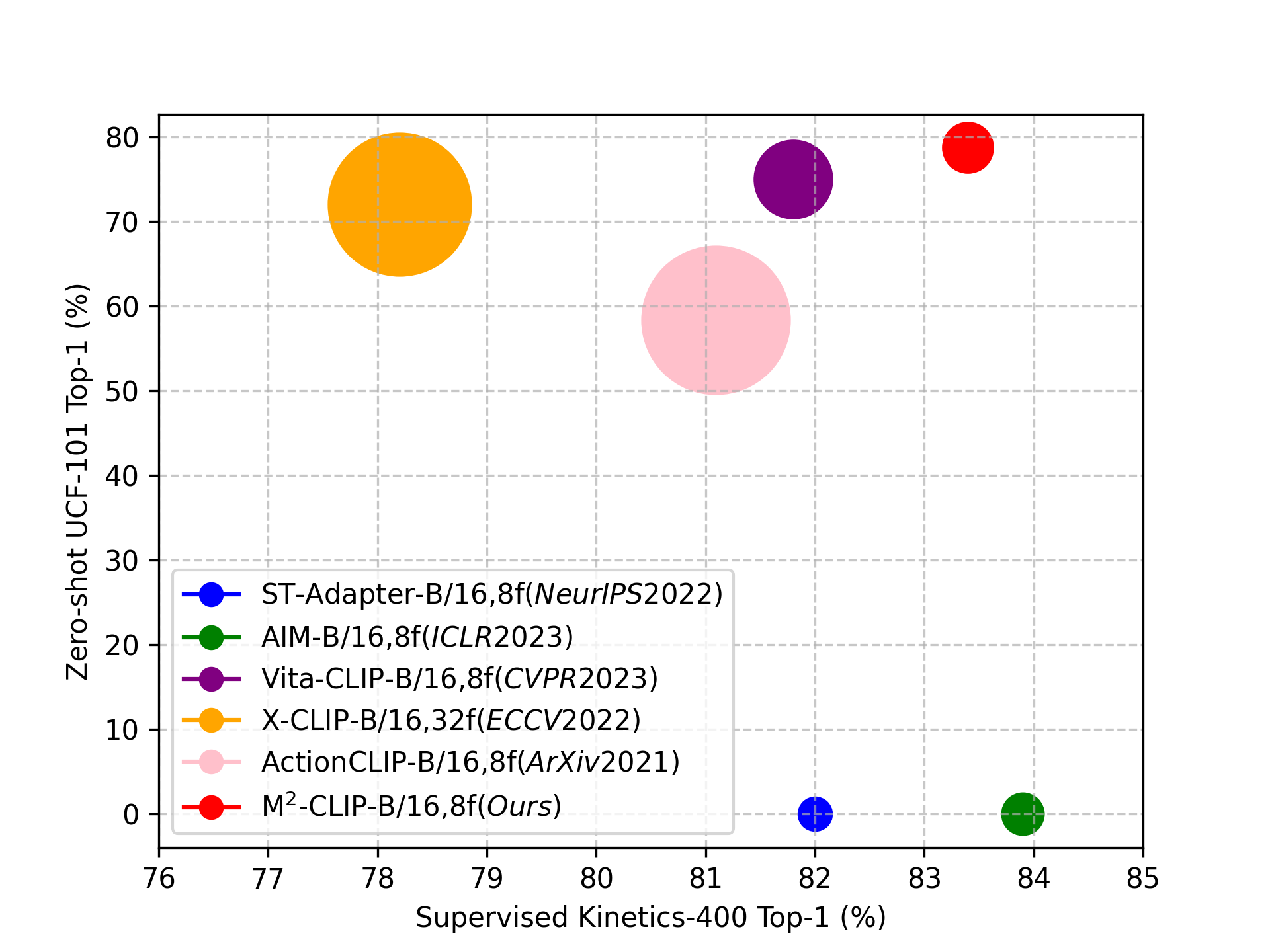}
    \small \caption{\textbf{Performance Comparison:} Zero-shot \textit{vs} supervised accuracy. The circle area represents the number of tunable parameters, where models with better performance are positioned towards the right and upper side, with a small circle area. Our \name achieves the best zero-shot performance  with very few tunable parameters.}
    \label{fig:performance}
    \end{figure}
   Over the past few years, there has been a remarkable surge of large-scale vision-language pre-trained models (VLM) like CLIP~\cite{clip}, ALIGN~\cite{align}, and Florence~\cite{florence}. As a result, researchers have actively delved into methods to effectively adapt these large models to their specific domains. In this paper, we focus on transferring the influential CLIP model to the domain of video action recognition, emphasizing its crucial role in driving advancements in this field.

   Undoubtedly, transferring knowledge from the powerful CLIP holds great promise due to its robust representation capability and impressive generalization performance. The most intuitive approach is to directly add temporal modeling to CLIP's image encoder and then finetune the entire network~\cite{action-clip,ILA,xclip}. However, finetuning comes with a high computational cost and may potentially impact the original generalization capabilities of CLIP.
    With the emergence of PEFT, researchers have begun to explore freezing the original CLIP parameters and introducing various adapters~\cite{stan,dualpath} or prompts~\cite{vita-clip,video-coop}, only training the newly added parameters. Notably, PEFT has motivated a reevaluation of the traditional unimodal video classification framework. By directly utilizing CLIP's visual branch in conjunction with added adapters, coupled with a Linear classification layer at the end, these approaches have demonstrated impressive results in supervised scenarios~\cite{frozen-clip,st-adapter,aim,dualpath,s-vit}. However, it is worth noting that excluding the text branch in these approaches leads to the loss of CLIP's generalization capabilities, which are among the fundamental attractions of the CLIP itself.

    \begin{figure}[t]
    \centering
    \includegraphics[width=.45\textwidth]{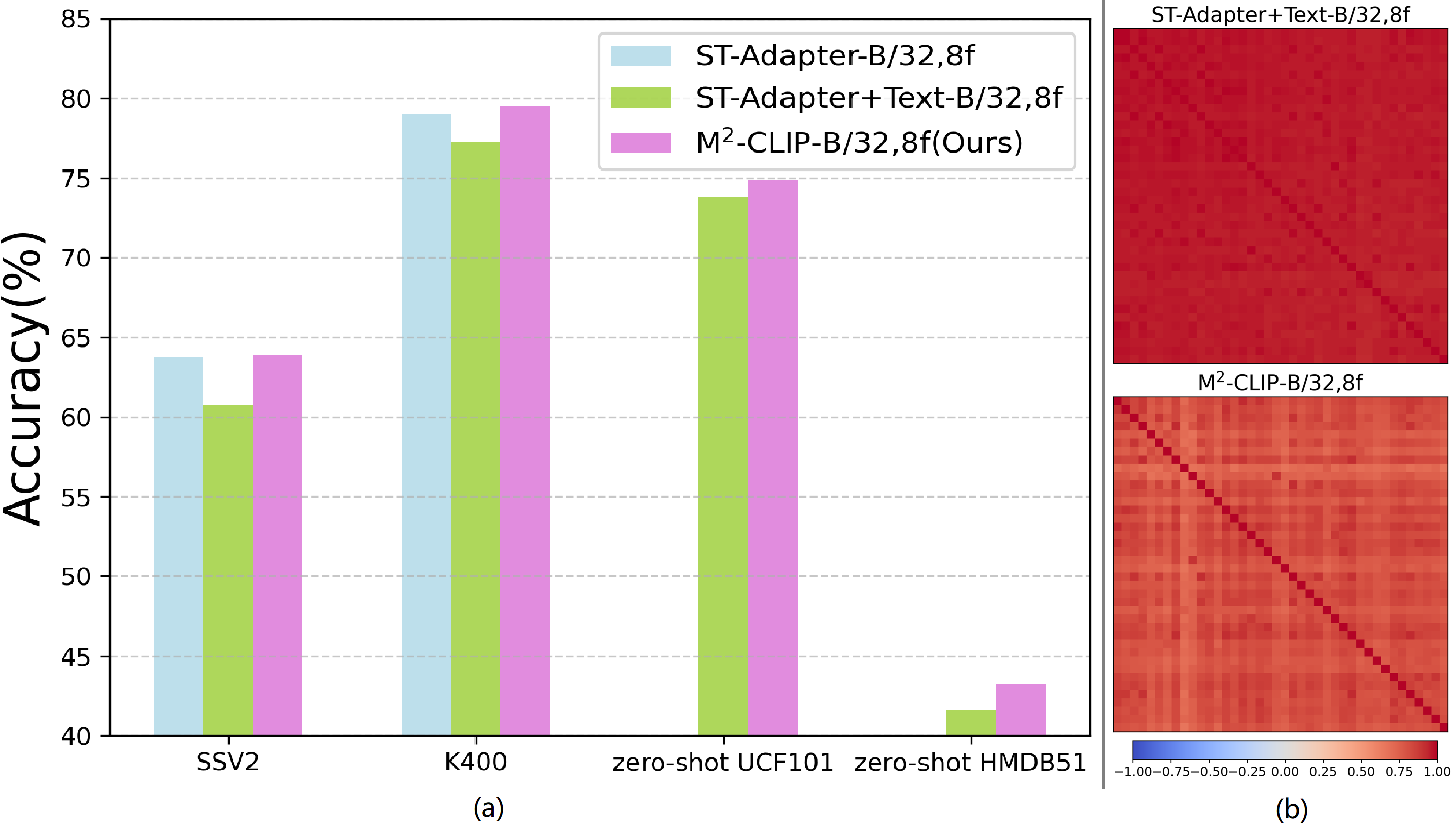}
    \caption{Analysis of transferring a unimodal framework into a multimodal one. (a) Performance comparison. Note that ST-Adapter is not able to zero-shot transferring, thus having no results in zero-shot UCF101 and HMDB51. (b) Inter-class correlation maps of the top 40 correlated SSv2 label features of ST-Adapter+text \textit{vs} the corresponding 40 label features of our method. The redder the color, the stronger the feature coupling. Our \name ultimately improved the performance on the four datasets and significantly reduced the correlation among the features of different labels.}
    \label{fig:intro}
    \end{figure}

    PEFT can also be applied to multimodal CLIP transfer frameworks, directly affecting the visual branch~\cite{stan} or the text branch~\cite{video-coop}, or even both simultaneously~\cite{vita-clip}. It significantly improves efficiency and reduces the number of learnable parameters. However, freezing the multimodal backbone causes a drop in supervised accuracy, leaving a gap compared to the performance of the unimodal frameworks, even when incorporating strong unimodal adapters. We experiment to validate this observation further, as shown in the left of Fig.~\ref{fig:intro}. Using ST-Adapter~\cite{st-adapter} as a representative of the unimodal frameworks, we introduce CLIP's text branch to transform ST-Adapter into a multimodal framework. Just as anticipated, by freezing the CLIP parameters while learning the adapters, we have indeed observed a noticeable decrease in supervised performance. The reason is that the text branch of CLIP lacks sufficient discriminative features, particularly for action verbs, as shown in the right of Fig.~\ref{fig:intro}. Then, the contrastive learning loss of CLIP itself makes it hard to learn discriminative features for videos when training with relatively small datasets compared with its original training set, especially when textual data is scarce.

    To mitigate the performance degradation while ensuring generalization, we propose a new multimodal, multi-task CLIP transfer framework, dubbed as \nameo. Firstly,  we focus on multimodal adapting to construct stronger architectures, adding adapters to both the text and visual branches. Specifically, to better represent the temporal information of videos, we design a novel TED-Adapter, capable of simultaneously integrating global temporal enhancement and local temporal difference modeling. In addition, we introduce a kind of naive adapter to the text branch to capture additional semantic information related to action labels, which significantly improves the first issue. Secondly, we devise a multi-task decoder for tapping into more substantial learning potential. The decoder consists of four components. (a) The first is the original contrastive learning head, which aims to align the pairwise video-text representations. (b) The second head is a cross-modal classification head, which can highlight the discriminative capabilities of cross-modal features. (c) Thirdly, we design a cross-modal masked language modeling head at the final layer of the text branch, promoting the focus of visual features on verbs for recognition. (d) Lastly, we incorporate a visual feature classifier at the end of the visual branch to facilitate the distinction of visual features across diverse categories.    
    
    In summary, our contributions are threefold: 1) We propose a novel multi-modal, multi-task adapting framework to transfer the powerful
    CLIP to video action recognition tasks. This method achieves strong supervised performance while ensuring state-of-the-art zero-shot transferability as shown in Fig.\ref{fig:performance}. 
    2) We design a new visual TED-adapter that performs Temporal Enhancement and Difference modeling to enhance the representation capabilities of the video encoder. Simultaneously, we introduce the adapters for the text encoder to make the label representation learnable and adjustable. 3) We introduce a multi-task decoder to improve the learning capability of the whole framework, adeptly achieving a balance between supervised performance and generalization.

\section{Related Works}
\subsection{Full Finetuning Video Action Recognition} 
 Early action recognition algorithms mostly relied on end-to-end finetuning of models pretrained on ImageNet~\cite{deng2009imagenet} with 2D CNNs~\cite{wang2016temporal,lin2019tsm,jiang2019stm}, 3D CNNs~\cite{feichtenhofer2019slowfast,feichtenhofer2020x3d} and Transformers~\cite{arnab2021vivit,li2021mvitv2,liu2021video,timesformer2021}. ImageNet pretrained models were primarily used to initialize these networks' backbones and expand to initialize some 3D convolutions~\cite{uniformerv2,carreira2017i3d}. Recently, the advent of large-scale image-language pretrained models like CLIP brought about significant changes. Given CLIP's strong performance on image-related tasks and remarkable generalization, researchers began full-finetuning from image to video based on CLIP. ActionCLIP~\cite{action-clip} was an early work that transferred CLIP to video action recognition by adding a temporal modeling module to CLIP's image branch, achieving competitive performance while keeping video generalization. X-CLIP~\cite{xclip} proposed frame-level temporal attention to reduce computation. These methods have achieved impressive results, but all require full finetuning on video data, making the training cost unaffordable for most researchers and practitioners.  
\subsection{PEFT in Video Action Recognition}
The PEFT technique initially emerged in the NLP field~\cite{houlsby2019parameter} to address the challenges of full finetuning for large-scale language models. In video action recognition, this technique has also become a research hotspot in recent years~\cite{video-coop,frozen-clip,aim,xing2023multimodal,stan}. EVL~\cite{frozen-clip} first proposed to leverage frozen CLIP image features with a lightweight spatiotemporal Transformer decoder to enhance video recognition tasks. ST-Adapter~\cite{st-adapter} introduced a parameter-efficient spatiotemporal adapter, which effectively harnesses the power of CLIP's image models for video understanding. AIM~\cite{aim} presented spatial, temporal, and joint adaptations to finetune pretrained image transformer models. These cost-effective methods have achieved SOTA performance, but they are all single-modal transfers, neglecting the text branch and thereby losing CLIP's generalization ability. Vita-CLIP~\cite{vita-clip} attempted to address this issue by adding prompts to both branches for transfer. However, we observed that its performance on temporally strong-correlated datasets was suboptimal, and their additional summary attention layers introduced in its visual branch increased the number of learnable parameters. In this work, we aim to apply PEFT to multi-modal frameworks to ensure competitive supervised performance with minimal increase in learnable parameters while maintaining strong generalization capabilities.

\section{Method}
\begin{figure*}[]
    \centering
    \includegraphics[width=0.96\textwidth]{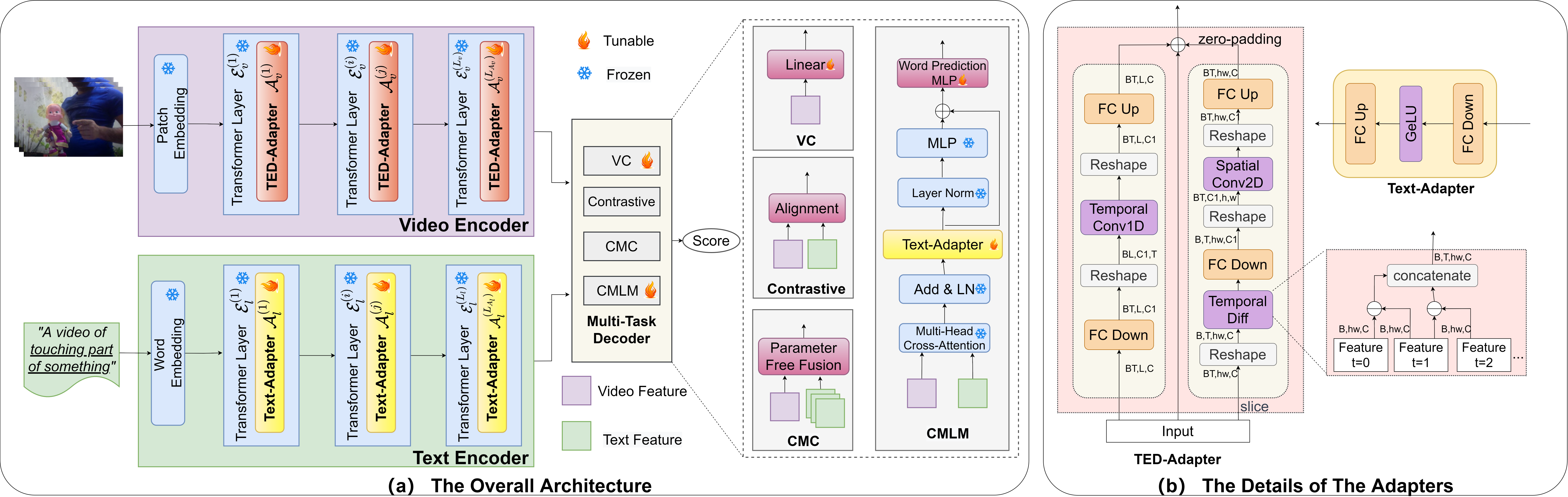}
    \caption{(a) Overview of \nameo: An example of integrating an adapter into each transformer layer is illustrated. \name consists of a video encoder, a text encoder and a multi-task decoder, where the backbones of the two encoders are frozen and assisted by the proposed trainable TED-Adapter and Text-Adapter. The multi-task decoder has four different heads that utilize multi-task constraints to improve the joint representation of the entire multimodal framework. (b) Detailed Structure of proposed adapters, where $L=1+M$ and $h \times w=M$. }
    \label{fig:overview}
\end{figure*}
\subsection{Architecture Overview}
As illustrated in Fig.~\ref{fig:overview}a, our framework comprises three key components: a video encoder, a text encoder, and a multi-task decoder. In this section, we will introduce the overview of the whole architecture and leave the details of the proposed multimodal adapters and the multi-task decoder in the following two sections.

Formally, the input to the framework is given as a video $\textbf{V}\in \mathbb{R}^{T\times H \times W \times 3}$ of spatial size $H \times W$ with $T$ sampled frames, and a text label $\textbf{y}$ from a predefined label set $\mathcal{Y}$.

\noindent\textbf{Video Encoder:} $\textbf{E}_v$ consists of $L_v$ transformer layers $\{\mathcal{E}_v^{(i)}\}_{i=1}^{L_v}$ and the proposed corresponding visual TED-Adapters $\{\mathcal{A}_v^{(j)}\}_{j=1}^{L_{A_v}}$. The $t$-th frame of the input divided into non-overlapping patches $\{\textbf{P}_{t,i}\}_{i=1}^{M}\in \mathbb{R}^{P^2 \times 3}$, $M=HW/P^2$. Then they are then projected into patch embeddings $\textbf{X}_{v,t} \in \mathbb{R}^{M \times d_v}$, prepended with a learnable class token $\textbf{C}_t$ and added with a positional encoding $\textbf{e}_v$. Mathematically, the frame-level input is constructed as:
\begin{equation}
[\textbf{c}_t^{(0)}, {\bf{z}}^{(0)}_{v,t}] = [\textbf{C}_t, \textbf{X}_{v,t}] + {\textbf{e}_v}.
\end{equation}

If we place the visual adapter before every transformer layer, the input will be sequentially processed as,
\begin{equation}
[\textbf{c}_t^{(i)}, {\bf{z}}^{(i)}_{v,t}] = \mathcal{E}_v^{(i)}(\mathcal{A}_v^{(i)}([\textbf{c}_t^{(i-1)}, {\bf{z}}^{(i-1)}_{v,t}]))  ~~~i= 1, 2, \cdots, L_v.
\end{equation}
To obtain the final video representation $\textbf{v}$, the class tokens $\text{c}_t^{(L_v)}$ of the last transformer layer is projected to a common video-language (VL) space by $\textbf{v}_t=\textbf{h}_v({\text{c}}_t^{(L_v)})$, and averaged along the temporal dimension,
\begin{equation}
\textbf{v} = \texttt{AvgPool}([{\bf{v}}_1, \cdots, {\bf{v}}_T]) \ \ \ \ \ ~~~~ \textbf{v} \in \mathbb{R}^{d_{vl}}.
\end{equation}
\noindent\textbf{Language Encoder:} Similarly, $\textbf{E}_l$ consists of $L_l$ transformer layers $\{\mathcal{E}_l^{(i)}\}_{i=1}^{L_l}$ and its corresponding text adapter $\{\mathcal{A}_l^{(j)}\}_{j=1}^{L_{A_l}}$. The input words are tokenized and projected into word embeddings $\textbf{X}_{l} \in \mathbb{R}^{N \times d_l}$, where $N$ is the text length. The input to the encoder is constructed as:
\begin{equation}
{\bf{z}}^{(0)}_{l} = \textbf{X}_{l} + {\textbf{e}_l}.
\end{equation}
Taking the example of inserting the text adapter before each transformer layer, the feature of each layer is obtained as:
\begin{equation}
{\bf{z}}^{(i)}_{l} = \mathcal{E}_l^{(i)}(\mathcal{A}_l^{(i)}({\bf{z}}^{(i-1)}_{l}))  ~~~i= 1, 2, \cdots, L_l.
\end{equation}
The final VL space text representation $\textbf{w} \in \mathbb{R}^{d_{vl}}$ of the label $\textbf{y}$ is obtained by $\textbf{w}=\textbf{h}_l({\bf{z}}^{(L_l)}_{l,N})$, where ${\bf{z}}^{(L_l)}_{l,N}$ is the last token of ${\bf{z}}^{(L_l)}_{l}$ and $\textbf{h}_l$ is a projection layer.

\noindent\textbf{Decoder:} 
Once the output features from the two encoders are obtained, they are fed into our specially designed multi-task decoder. In the training process, the role of the decoder is to impose constraints on the feature representations generated by the encoders, facilitating semantic alignment between the two modalities and enabling differentiation between features of different categories. Once a model completes its training, the decoder is versatile, capable of generating classification scores for supervised learning and conducting zero-shot classification. The detailed design of the decoder's structure will be elaborated in the following section.
\subsection{Visual and Textual Adapters}
To better transfer CLIP to this task and enhance the semantic representation of action verbs in the labels, we introduce adapters for both the visual and text branches to improve their respective representation capabilities.

\noindent \textbf{Video TED-Adapter:} Adapting CLIP's image branch to the video branch requires additional temporal modeling modules, which can be approached from two perspectives, global temporal enhancement and local temporal difference modeling. The former is the intuitive global temporal aggregation referred to as spatiotemporal features~\cite{lin2019tsm,feichtenhofer2019slowfast}, where temporal attentions or temporal convolutions are applied to multiple frames' features to aggregate the similar action subject. It has been extensively explored in CLIP's transfer~\cite{st-adapter,aim,stan}. The latter is short-term frame-wise feature difference learning, which seeks to capture the local motion patterns and dynamics between adjacent frames. This kind of feature has been mentioned in earlier computation-efficient convolutional algorithms~\cite{jiang2019stm,stm,tea} but remains unexplored in the context of CLIP's transfer. To explore both the two kinds of temporal modeling in a unified structure, we design a novel TED-Adapter, which learns the \textbf{T}emporal \textbf{E}nhancements and temporal \textbf{D}ifferences in the meanwhile.

As shown in Fig.~\ref{fig:overview}b, we first adopt a 1D temporal convolution for temporal feature enhancement. For the input of a TED-Adapter layer including the class token and patch tokens $\textbf{Z}=\{[\textbf{c}_t, {\textbf{z}}_{v,t}]\}_{t=1}^{T}\in \mathbb{R}^{ T\times (1+M) \times d_v }$, we perform the following operations:
\begin{equation}
\textbf{Z}_{E} = \texttt{Conv1D}(\textbf{Z} \textbf{W}_{dn})\textbf{W}_{up},
\end{equation}
where $\textbf{W}_{dn}$ and $\textbf{W}_{up}$ are the down-projection and up-projection weights. \texttt{Conv1D} represents the 1D-convolution for spatiotemporal modeling operating on the temporal dimension. Note that the reshape operations are omitted in this section for simplicity but are shown in Fig.~\ref{fig:overview}b.

Next, for the temporal difference modeling, we subtract the previous frame's feature from the current frame and then employ a 2D spatial convolution to learn useful information from the adjacent feature differences automatically. Formally, when given the input patch tokens $\textbf{z}_{v,t}$ the of the $t$-th frame,
\begin{equation}
\textbf{z}_{D,t} = \texttt{Conv2D}((\textbf{z}_{v,t}-\textbf{z}_{v,t-1}) \textbf{W}_{dn})\textbf{W}_{up},
\end{equation}
where \texttt{Conv2D} represents the 2D spatial convolution. For the first frame, we set its feature differences to zeros.

Finally, the output of TED-Adapter can be obtained by fusing the two kinds of temporal features together. Moreover, a residual summation is applied to preserve the information in the input:
\begin{equation}
\mathcal{A}_v([\textbf{c}_t, {\textbf{z}}_{v,t}]) = \textbf{Z}_{E} + \textbf{Z}_{D} + \textbf{Z},
\end{equation}
where $\textbf{Z}_{D}=\{[\textbf{O}, \textbf{z}_{D,t}]\}_{t=1}^{T}$, and ${\textbf{O}}$ is a zero matrix which has the same shape as $\textbf{c}_t$.

The TED-Adapter is simply placed before the Multi-Head Self-Attention (MHSA) by default unless otherwise specified. By incorporating the temporal enhancement and temporal difference operations, the proposed TED-Adapter can capture spatiotemporal features and local finer motion patterns, which are both crucial for this task.

\noindent \textbf{Text Adapter:} In action recognition, the textual labels describing actions are often short and succinct, emphasizing the actions themselves, such as ``unfolding something" and ``hurdling". However, we observed that CLIP's text encoder alone might not effectively distinguish such label text features, as shown in Fig.~\ref{fig:intro}. To address this, we introduce adapters to the text branch to learn better semantic representations for the action labels. We directly utilized the basic adapter~\cite{houlsby2019parameter} structure here as shown in Fig.~\ref{fig:overview}b. Specifically, given the input text tokens of a text adapter layer ${\textbf{z}}_l$, we perform the text adapter like:
\begin{equation}\label{eq:textadapter}
\mathcal{A}_l(\textbf{z}_{l}) = \textbf{z}_{l} +\texttt{Act}({\textbf{z}}_l \textbf{W}_{dn})\textbf{W}_{up},
\end{equation}
where $\texttt{Act}$ means a non-linear activation function and we use GeLU here. 

The text adapters are inserted before the Feed-Forward Networks (FFN) of the transformer layer by default. By incorporating the text adapter, the model can enhance its understanding of the action labels, capturing more discriminative semantic information. This allows for improved alignment between the textual and visual representations, resulting in more accurate and effective video action recognition. 

\subsection{Multi-Task Decoder}
As previously described, we observed that when utilizing CLIP's multimodal framework, relying solely on contrastive learning did not perform as well as the equivalently configured unimodal framework. To address this, we propose a multi-task decoder equipped with four distinct learning tasks, each corresponding to a separate head, as shown in the right part of Fig.~\ref{fig:overview}a. This approach aims to leverage multiple task constraints to improve the joint representation power of the multimodal framework.

\noindent \textbf{Multimodal Contrastive Learning Head (Contrastive).} This is the original training objective of CLIP. To pull the pairwise video representation $\textbf{v}$ and label representation $\textbf{w}$ close to each other, symmetric similarities are defined between the two modalities:
\begin{equation}\label{contrastive}
	\begin{aligned}
		p_{i}^{\textbf{\textbf{V}}2\textbf{\textbf{y}}}(\textbf{\textbf{V}})&=\frac{\mathrm{exp}(\texttt{cos}(\textbf{\textbf{v}},\textbf{w}_{i})/\tau)}{\sum_{j=1}^{B}\mathrm{exp}(\texttt{cos}(\textbf{\textbf{v}},\textbf{w}_{j})/\tau)},\\
		p_{i}^{\textbf{\textbf{y}}2\textbf{\textbf{V}}}(\textbf{\textbf{y}})&=\frac{\mathrm{exp}(\texttt{cos}(\textbf{\textbf{w}},\textbf{v}_{i})/\tau)}{\sum_{j=1}^{B}\mathrm{exp}(\texttt{cos}(\textbf{\textbf{w}},\textbf{v}_{j})/\tau)},
	\end{aligned}
\end{equation}
where $\texttt{cos}$ means cosine similarity, $\tau$ is a temperature parameter and $B$ is the number of training pairs. The ground-truth is defined as 0 for negative pairs and 1 for positive pairs. We use Kullback-Leibler divergence as the video-text contrastive loss to optimize this head as ActionCLIP.

When a model is trained, it will be ready for zero-shot classification. In practice, the text input can be prompted like ``\texttt{a video of }$<\hat{\textbf{y}}>$", where $\hat{\textbf{y}}$ is a category name of $C$ classes. The process of predicting $\hat{\textbf{y}}$ of a certain video $\textbf{V}$ is to find the highest similarity score calculated by:
\begin{equation}\label{zsinfer}
p(\hat{\textbf{y}}|\textbf{V}) = \frac{\text{exp}(\texttt{cos}(\textbf{v}, \textbf{w}_{\hat{y}})/\tau)}{\sum_{i=1}^{C}\text{exp}(\texttt{cos}(\textbf{v}, \textbf{w}_{i}))}.
\end{equation} 

\noindent \textbf{Cross-Modal Classification Head (CMC).} 
Since the action labels are predefined within a given set ($C$ classes), we can compute the complete label feature set for each iteration, enabling us to carry out cross-modal feature classification. In this work, we employed a straightforward parameter-free cross-modal fusion approach, which directly computes cosine similarity using the Eq.~(\ref{zsinfer}). Note that it differs from Eq.~(\ref{contrastive}), which considers video-text matching within a training batch and can not cover all the action labels' representations. After obtaining these similarities, the goal is to ensure that a video's representation is similar to the textual representation of its corresponding label rather than the textual representations of other categories. To achieve this, we ingeniously transform the problem into a 1-in-$C$ classification task and add a classification constraint to the cross-modal similarity scores with the cross-entropy loss. 

\noindent \textbf{Cross-Modal Masked Language Modeling Head (CMLM).} 
Unlike the original CLIP, which primarily deals with image-text paired data, our action labels predominantly focus on verbs. To enhance CLIP's text branch for better representation of action-related words and help the learning of the text adapters, we introduce an additional CMLM head, which urges the text branch to predict masked words from the other text and video tokens. Specifically, given the framewise video features $[{\bf{v}}_1, \cdots, {\bf{v}}_T]$ and the text features ${\bf{z}}^{(L_l)}_{l}$, we perform a cross-attention operation to obtain cross-modal features. Due to the limited amount of textual data, directly learning the parameters of this attention layer can be challenging. Our approach to addressing this is to initialize the parameters of this attention layer using the parameters of the final transformer layer in the text branch and then freeze these parameters and add a text adapter of Eq.~(\ref{eq:textadapter}) before the FFN of the transformer layer. Then we only learn the parameters of this text adapter. The process can be presented as:
\begin{equation}
\begin{aligned}
     \textbf{w}^* &= {\bf{z}}^{(L_l)}_{l}+\texttt{CA}(\texttt{LN}({\bf{z}}^{(L_l)}_{l}),\texttt{LN}([{\bf{v}}_1, \cdots, 
    {\bf{v}}_T])) ,\\
     \hat{\textbf{w}} &= \mathcal{A}_l(w^*),\\
     \textbf{w}_{m} &=  \hat{\textbf{w}} +  \texttt{MLP}(\texttt{LN}(\hat{\textbf{w}})),
\end{aligned}   
\end{equation}
where \texttt{CA}, \texttt{LN} and \texttt{MLP} indicate the cross attention layer, layer norm and a MLP layer, respectively. Then, we attach a BERT MLM head~\cite{bert} to predict the masked words with a cross-entropy loss, as shown in Fig.~\ref{fig:overview}a.

\noindent \textbf{Visual Classification Head (VC).}
Furthermore, we introduced a straightforward classification head to the video branch to enhance the distinction between different categories in video features. Given video features $\textbf{v}$, we directly appended a Linear layer for classification, training with cross-entropy loss. Importantly, with the inclusion of this classification head, we can directly use its output for supervised classification tasks. For zero-shot experiments, we still employ Eq. (\ref{zsinfer}). The addition of this classification head enables the model to learn to discriminate the video features between different action categories more effectively. 

In summary, by introducing these four learning tasks, we tap into a richer set of supervisory signals, guiding the model to better align the visual and textual modalities while simultaneously capturing various aspects of semantic information. This multi-task approach not only mitigates the performance disparity of supervised learning but also preserves CLIP's remarkable generalization capabilities.
\begin{table*}[ht]

\centering
\small
\renewcommand\arraystretch{1.1}
\scalebox{0.73}{
\begin{tabular}{lcccccccc}
\rowcolor{LightCyan}
    \toprule
    \textbf{Method }                                                          & \textbf{Pre-training}    &\textbf{Tunable Param} & \textbf{\#Frames}     & \textbf{Top-1(\%)} &\textbf{ Top-5(\%)} & \textbf{GFLOPs}    & \textbf{Zero-shot} \\
    \midrule
    \multicolumn{7}{l}{\textit{Full Finetuning}} \\
     Swin-B~\textit{(CVPR'22)}~\cite{liu2021video}                    & IN-21k      &88        & 32 × 4 × 3     & 82.7  & 95.5  & 282       & \xmark \\
    MViTv2-B~\textit{(CVPR'22)}~\cite{li2021mvitv2}                  & \xmark            &52       & 32 ×  5 × 1     & 82.9  & 95.7  & 225       & \xmark \\
    Uniformer V2-B/16~\textit{(ICLR'23)}~\cite{uniformerv2}        & CLIP-400M &115 & 8 × 3 × 4               & 85.6                    & 97.0                    & 154    \\
     ActionCLIP-B/16~\textit{(arXiv'21)}~\cite{action-clip}           & CLIP-400M       &142 & 32 × 10 × 3    & 83.8  & 96.2  & 563       & \cmark \\
    X-CLIP-B/16~\textit{(ECCV'22)}~\cite{xclip}                      & CLIP-400M       &132  & 16 × 4 × 3     & 84.7  & 96.8  & 287       & \cmark \\
    BIKE-L/14~\textit{(CVPR'23)}~\cite{bike}                         & CLIP-400M      &230  & 16 × 4 × 3     & \textbf{88.1}  & 97.9  & 830       & \cmark \\ 
    S-ViT-B/16~\textit{(CVPR'23)}~\cite{s-vit}                       & CLIP-400M      &-  & 16 × 3 × 4     & 84.7  & 96.8  & 340      & \xmark \\ 
    ILA-ViT-L/14~\textit{(ICCV'23)}~\cite{ILA}                       & CLIP-400M      &-  & 8 × 4 × 3     & 88.0  & \textbf{98.1}  & 673       & \cmark \\ 
    \midrule
     \multicolumn{7}{l}{\textit{PEFT: unimodal visual framework (frozen CLIP)}} \\
    EVL-B/16~\textit{(ECCV'22)}~\cite{frozen-clip}                   & CLIP-400M        &86 & 8 × 1 × 3     & 82.9  & -     & 444       & \xmark \\
    ST-Adapter-B/16~\textit{(NeurIPS'22)}~\cite{st-adapter}          & CLIP-400M        &7 & 8 × 1 × 3     & 82.0  & 95.7  & 148       & \xmark \\
    ST-Adapter-B/16~\textit{(NeurIPS'22)}~\cite{st-adapter}          & CLIP-400M        &7 & 32 × 1 × 3     & 82.7  & 96.2  & 607       & \xmark \\
     AIM-B/16~\textit{(ICLR'23)}~\cite{aim}                           & CLIP-400M       &11  & 8 × 1 × 3     & 83.9  & 96.3  & 202       & \xmark \\
    AIM-B/16~\textit{(ICLR'23)}~\cite{aim}                           & CLIP-400M       &11  & 32 × 1 × 3     & 84.7  & 96.7  & 809       & \xmark \\
     DUALPATH-B/16~\textit{(CVPR'23)}~\cite{dualpath}                           & CLIP-400M       &10  & 32 × 1 × 3     & \textbf{85.4}  & \textbf{97.1}  & 237       & \xmark \\
    \midrule
    \multicolumn{7}{l}{\textit{PEFT: multimodal framework (frozen CLIP)}} \\
    
   STAN-conv-B/16~\textit{(CVPR'23)}~\cite{stan}                              & CLIP-400M     &-   & 8 × 1 × 3     & 83.1  & 96.0  & 238        & \cmark \\
   Vita-CLIP B/16~\textit{(CVPR'23)}~\cite{vita-clip}                              & CLIP-400M     &39   & 8 × 4 × 3     & 81.8  & 96.0  & 97        & \cmark \\
    Vita-CLIP B/16~\textit{(CVPR'23)}~\cite{vita-clip}                             & CLIP-400M      &39  & 16 × 4 × 3     & 82.9 & 96.3 & 190       & \cmark \\
     \rowcolor{LightCyan}  \textbf{M$^2$-CLIP}-B/16         & CLIP-400M       &16  & 8 × 4 × 3     & 83.4  & 96.3  & 214      & \cmark \\
  \rowcolor{LightCyan}  \textbf{M$^2$-CLIP}-B/16         & CLIP-400M       &16  & 16 × 4 × 3     & 83.7  & 96.7  & 422     & \cmark \\
  \rowcolor{LightCyan}  \textbf{M$^2$-CLIP}-B/16        & CLIP-400M       &16  & 32 × 4 × 3     & \textbf{84.1}  & \textbf{96.8}  & 842       & \cmark \\
    
    \bottomrule
\end{tabular}}
\caption{Performance comparison on K400. The per-view GFLOPs is reported. \#Frame means frames$\times$crops$\times$clips.}
\label{tab:sup_k400}
\end{table*}

\begin{table}[t!]\small
    \begin{center}
    \renewcommand\arraystretch{1.2}
    \scalebox{0.72}{
    \begin{tabular}{lccc}
    \rowcolor{LightCyan}
    \toprule
    \textbf{Model}       &  \textbf{\#Frames}          & \textbf{Top-1(\%)} & \textbf{Top-5(\%)} \\ \hline
        \multicolumn{4}{l}{\textit{Full Finetuning}} \\
    ViViT-L~\cite{arnab2021vivit}                          & 16×1×3              & 65.4                    & 89.8   \\
     Mformer-B~\cite{patrick2021keeping}                       & 16×1×3                & 66.5                    & 90.1        \\
    MViTv2-B~\cite{li2021mvitv2}                                   & 32×1×3               & 70.5                    & 92.7     \\
        ILA-ViT-B/16~\cite{ILA}                       & 8×4×3               & 65.0                    & 89.2        \\
    ILA-ViT-B/16~\cite{ILA}                        & 16×4×3                & 66.8                    & 90.3                         \\
    Uniformer V2-B/16~\cite{uniformerv2}       & 32×1×3               & \textbf{70.7}                    & \textbf{93.2}    \\
    S-ViT-B/16~\cite{s-vit}                & 16×2×3      & 69.3  & 92.1     \\ 
    \midrule
        \multicolumn{4}{l}{\textit{PEFT: unimodal visual framework (frozen CLIP)}} \\
    ST-Adapter-B/16~\cite{st-adapter}          & 8×1×3     & 67.1  & 91.2 \\
    ST-Adapter-B/16~\cite{st-adapter}        & 32×1×3     & 69.5  & 92.6  \\
    EVL-ViT-B/16~\cite{frozen-clip}                        & 16×1×3               & 61.7                    & -                                               \\
    DUALPATH-B/16~\cite{dualpath}                          &32×1×3    & \textbf{70.3}  & \textbf{92.9}   \\
    
    AIM-ViT-B/16~\cite{aim}        & 8×1×3               & 66.4                    & 90.5                        \\
    AIM-ViT-B/16~\cite{aim}        & 32×1×3               & 69.1                    & 92.2                           \\
    
     \midrule
        \multicolumn{4}{l}{\textit{PEFT: multimodal framework (frozen CLIP)}} \\
    
    STAN-conv-B/16~\cite{stan}         &8×1×3       & 65.2  & 90.5   \\
     Vita-CLIP-B/16~\cite{vita-clip}        & 16×-     &48.7  & -      \\
       \rowcolor{LightCyan}   \textbf{M$^2$-CLIP}-B/16           &  8×1×3    & 66.9  &90.1   \\
        \rowcolor{LightCyan}    \textbf{M$^2$-CLIP}-B/16           & 32×1×3   & \textbf{69.1} & \textbf{91.8}  \\
    \bottomrule
    \end{tabular}
    }
    \end{center}
    \caption{Performance comparison on SSv2.
    }
    \label{tab:sup_ssv2}
    \end{table}

\section{Experiments}
\subsection{Experimental Setup}
We evaluate our \name for supervised learning in two primary datasets: Kinetics-400 (K400)~\cite{k400} and Something-Something-V2 (SSv2)~\cite{goyal2017ssv2}. For the generalization evaluation, we test our model on UCF101~\cite{soomro2012ucf101} and HMDB51~\cite{kuehne2011hmdb}. We employ ViT-B/16 based CLIP as our backbone and use a sparse frame sampling strategy with 8, 16, or 32 frames during training and inference. %
\subsection{Fully-Supervised Experiments}
We present our results of K400 and SSV2 in Tab.~\ref{tab:sup_k400} and Tab.~\ref{tab:sup_ssv2}, respectively, comparing our approach with SOTAs trained under various transfer methods, including full finetuning, unimodal and multimodal PEFT from frozen CLIP.

On K400, our 8-frame \nameo-B/16 model surpasses models pretrained by ImageNet~\cite{deng2009imagenet}, achieving higher performance with fewer learnable parameters and computational requirements. Compared to end-to-end finetuned CLIP models with the same ViT-B/16 backbones, our approach demonstrates comparable results. With just 11\% of the adjustable parameters, we exceed ActionCLIP~\cite{action-clip}'s results (84.1\% vs. 83.8\%). Furthermore, our method is competitive with the latest approaches like X-CLIP~\cite{xclip} and S-ViT~\cite{s-vit}, yet with much fewer learnable parameters. In addition, while our results fall slightly short when compared to the leader performances achieved by BIKE~\cite{bike} and ILA~\cite{ILA}, it's important to note that they employed a much larger network architecture (ViT-L) and had 14 times the number of tunable parameters as our model.
Compared with the unimodal PEFT approaches, our method achieves comparable or even superior results. For instance, our 8-frame \nameo-B/16 model outperforms 8-frame ST-Adapter-B/16~\cite{st-adapter} by 1.4\%. It is worth noting that while unimodal methods exhibit high performance in supervised settings, they lack support for zero-shot generalization. In contrast, our approach achieves competitive results with them and demonstrates strong generalizations.
Lastly, compared with multimodal PEFT approaches, our method achieves superior results. Note that Vita-CLIP~\cite{vita-clip} is a multimodal Prompt-based method while we use adapters. It is evident that we achieve higher performance with only 41\% trainable parameters.

As for SSv2, our approach achieves comparable performance and even surpasses several full-finetuned methods with similar configurations, such as Mformer-B~\cite{patrick2021keeping} and ILA-ViT-B~\cite{ILA}, while utilizing fewer trainable parameters and computational resources. Compared with unimodal PEFT methods, our 8-frame \name model surpasses AIM-ViT-B/16~\cite{aim} and EVL-ViT-B/16~\cite{frozen-clip} in Top-1 performance and maintains a competitive position compared to other methods. In the domain of multimodal PEFT approaches, our method outperforms the recent method Vita-CLIP~\cite{vita-clip} by a large margin of over 18\%. The results demonstrate that our proposed multimodal adapters and the multi-task decoder are helpful strategies for efficient multimodal CLIP-based image-to-video knowledge transfer.

\subsection{Zero-shot Experiments}
\begin{table}[t!]
\centering
\small
\renewcommand\arraystretch{1.2}
\resizebox{1\columnwidth}{!}{
\begin{tabular}{lccc}
\rowcolor{LightCyan}
    \toprule
	Method                                                    & HMDB51(\%)           & UCF101(\%)     &CLIP-FT \\
    \midrule
	ResT-101~\textit{(CVPR'22)}~\cite{rest}         & 46.7    & 34.4 & N/A\\
    ActionCLIP~\textit{(arXiv'21)}~\cite{action-clip}                   & 40.8 $\pm$ 5.4    & 58.3 $\pm$ 3.4 & \cmark\\
	X-CLIP-B/16~\textit{(ECCV'22)}~\cite{xclip}               & 44.6 $\pm$ 5.2    & 72.0 $\pm$ 2.3 & \cmark \\
 A5~\textit{(ECCV'22)}~\cite{video-coop}                   & 44.3 $\pm$ 2.2    & 69.3 $\pm$ 4.2 & \xmark\\
       CoOp~\textit{(IJCV'22)}~\cite{coop}                      &  - &66.6   & \xmark \\
    Co-CoOp~\textit{(CVPR'22)}~\cite{cocoop}                &- &68.2   & \xmark \\
    MaPLe~\textit{(CVPR'23)}~\cite{maple}                &- &68.7   & \xmark \\
	Vita-CLIP-B/16~\textit{(CVPR'23)}~\cite{vita-clip}    & \textbf{48.6}  $\pm$ 0.6 & 75.0 $\pm$ 0.6 & \xmark\\
 \midrule
 \rowcolor{LightCyan} \textbf{M$^2$-CLIP}-B/16   & 47.1  $\pm$ 0.4 & \textbf{78.7} $\pm$ 1.2 & \xmark\\
    \bottomrule
\end{tabular}
}
\caption{Comparison for zero-shot performances on HMDB51 and UCF101. CLIP-FT means CLIP finetuning.}

\label{tab:zeroshot_hmdb_ucf}
\hfill
\end{table}
\begin{table}[t!]

\centering
\small
\scalebox{0.95}{\begin{tabular} {l|c}  
\rowcolor{Gray}
\toprule
Components & Top-1(\%) \\
\midrule
Baseline(CLIP zero-shot)  &  56.5\\
	+ TED-Adapter & 80.5  \\
 + Multimodal-Adapter &  81.4 \\
  + Multimodal-Adapter + Multi-Task Decoder & 83.4 \\
    \bottomrule
\end{tabular}
}
\caption{Ablations for Components.}
\label{tab:components}
\hfill

\end{table}
\captionsetup[subfigure]{font=small}
\begin{figure*}[ht]
    \centering
    \begin{minipage}[t]{0.32\textwidth}
    \centering
    \subfloat[]{
     \small
        \renewcommand{\arraystretch}{0.7}
        \scalebox{0.85}{
        \begin{tabular} {lc|cc|cc|c}
\rowcolor{Gray}
\toprule
	1-5 & 6-12 & TE &TD &Sequential &Parallel  & Top-1(\%) \\
    \midrule
    	\cmark & &\cmark &\cmark & &\cmark  & 80.4 \\
     & \cmark &\cmark  &\cmark  & &\cmark & 82.9  \\
     \midrule
   \cmark  & \cmark & \cmark &    & &  &83.1 \\
   \cmark  & \cmark &  &\cmark & &  &81.8  \\
   \midrule
    \cmark & \cmark & \cmark &\cmark &\cmark  &  & 83.3 \\
    \cmark & \cmark & \cmark &\cmark &  & \cmark & 83.4 \\
    \bottomrule
\end{tabular}}}
   \end{minipage}\hfill
\begin{minipage}[t]{0.32\textwidth}
    \centering
    \subfloat[]{
        \raisebox{-1.2cm}{\includegraphics[width=0.6\textwidth]{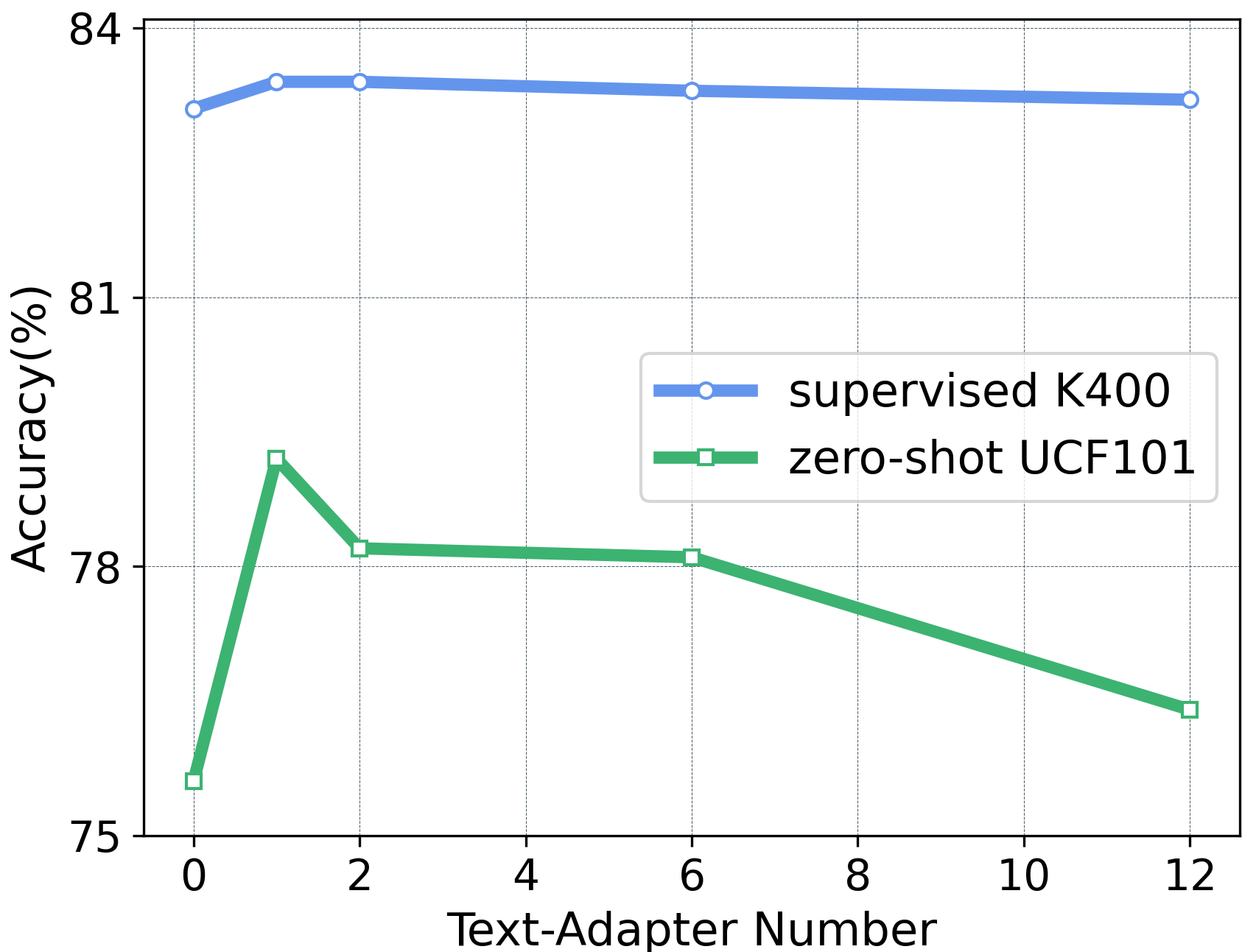}} 
    }
    \hspace{-1.5cm}
    \end{minipage}
   \begin{minipage}[t]{0.35\textwidth}
   \centering
    \subfloat[]{
        \renewcommand{\arraystretch}{1.2}
        \scalebox{0.81}{
       \begin{tabular} {ccccccc}
\rowcolor{Gray}
\toprule
	Contrastive & CMC & MLM &VC  & Top-1(\%) \\
    \midrule
     \cmark &  & &    & 81.4 \\
    	\cmark &\cmark & & &82.1  \\
   \cmark  & \cmark &\cmark  & & 82.4  \\
   \cmark  & \cmark & \cmark & \cmark   &83.4 \\
    \bottomrule
\end{tabular}}

    }
    
    \end{minipage}\hfill
    \caption{Ablation experiments for: (a) TED-Adapter, (b) Text-Adapter and (c) Multi-task decoder.}
    \label{fig:ablations}
\end{figure*}

In this experiment, we employ the \nameo-B/16 model pre-trained on K400, with 8 frames as input, for conducting generalization experiments on UCF101 and HMDB51. We use the outputs of the contrastive learning head for classification. It's worth noting that the model used for testing is consistent with the one mentioned in the supervised experiments in Tab.~\ref{tab:sup_k400}, as Vita-CLIP\cite{vita-clip}. 

Our approach demonstrates impressive generalization capabilities as shown in Tab.~\ref{tab:zeroshot_hmdb_ucf}. All methods in this table, except ResT~\cite{rest}, are transferred from CLIP. In comparison to methods that require full finetuning, our results on both datasets outperform them by a significant margin. For instance, we surpass X-CLIP~\cite{xclip} by 2.5\% on HMDB51 and 6.7\% on UCF101. Moreover, our method requires far fewer trainable parameters than these approaches. Among other methods that do not require full finetuning and are based on Prompts, our method also outperforms the majority. While we may have a 1.5\% lower performance than Vita-CLIP on HMDB51, our model uses only 41\% of the trainable parameters of them and has better supervised results. Furthermore, our accuracy on UCF101 is 2.7\% higher than theirs, making our approach the leader among all the methods on this dataset.

\subsection{Ablation and Analysis}

In this section, unless otherwise specified, we use ViT-B/16 as the backbone and 8 input frames in all ablation experiments on K400. \\
\noindent \textbf{Effectiveness of Components.} 
In Tab.~\ref{tab:components}, we first construct a CLIP frozen baseline and use its zero-shot performance as our baseline. Then, we gradually introduced our contributions based on this baseline. We have observed a significant performance boost when we introduce the learnable temporal modeling module, TED-Adapter, demonstrating its effectiveness. Subsequently, we add the Text-Adapter, which formulates the multimodal adapters with TED-Adapter to CLIP and further improves the performance. Finally, by adding the Multi-task decoder, we form the final \nameo, incorporating multiple learning objectives and resulting in a substantial improvement. In summary, our proposed multimodal adapter and multi-task decoder are both highly effective and modular components that can be easily integrated into CLIP transfer frameworks as plug-and-play modules.\\
\noindent \textbf{Ablations for Video TED-Adapter.} We next conducted ablation experiments on Video TED-Adapter in Fig.~\ref{fig:ablations}a. We use one TED-Adapter with bottleneck width 384 as ST-Adapter~\cite{st-adapter}. First, we attempt to add the TED-Adapter to the front half and the back half of the network. We have observed that although more TED-Adapters generally lead to better results, the benefits gained from adding them to deeper layers outweigh those from shallow layers. Secondly, as our TED-Adapter includes temporal enhancement (TE) and temporal difference (TD), we conducted separate experiments, revealing that the improvement from TE is more pronounced, but the combination of both yields the best performance. Finally, we found that the parallel addition of components is slightly better than the sequential one. \\
\noindent \textbf{Ablations for Text-Adapter.}
We progressively added a varying number of Text-Adapters from deep to shallow. We show the performance changes in Fig.~\ref{fig:ablations}b, along with the corresponding zero-shot transfer performance on UCF101. It can be observed that as the number of Text-Adapters increases, the model's performance first improves and then starts to decline slightly, with the best performance achieved when adding just one Text-Adapter. The zero-shot results exhibit a similar trend of variation. We believe the reason is that the text data contains only label information, and having too many Text-Adapters may lead the model to overfit on these labels, ultimately affecting overall performance and generalization. Therefore, our final model includes one Text-Adapter, balancing performance and generalization.\\
\noindent \textbf{Ablations for Multi-Task Decoder.} In Fig.~\ref{fig:ablations}c, we evaluate the impact of the individual heads in the decoder. It is evident that each head in our model contributes positively to the results. By incorporating CMC on top of the original CLIP's contrastive learning, we achieve additional 0.7\% improvements in performance, and notably, this enhancement is parameter-free. Furthermore, the inclusion of CMLM further boosted the results. Lastly, the addition of the VC head elevated the performance to 83.4\%. This comprehensive analysis demonstrates our multi-task decoder's effectiveness in enhancing the multimodal framework's overall learning.

\section{Conclusion}
In this paper, we introduced a novel multimodal, multi-task adapting approach that addresses the challenging task of transferring a large vision-language model, CLIP, to the domain of video action recognition. Our core innovation lies in integrating multimodal adapters and a multi-task decoder into the multimodal framework. The multimodal adapters, specifically designed for handling both visual and textual branches, enable our model to effectively leverage the rich information contained in these modalities and contribute to better feature extraction and understanding. The multi-task decoder, another key component, introduces diverse learning objectives, promoting performance by simultaneously addressing various tasks. Comprehensive experiments on various datasets showcase our method's remarkable zero-shot performance while maintaining promising supervised results with few tunable parameters.

\section*{Acknowledgments}
This work is funded by Collective Intelligence \& Collaboration Laboratory (Open Fund Project No.QXZ23012301).

	\bibliography{aaai24}

\end{document}